\documentclass{article} 
\usepackage[T1]{fontenc}
\usepackage{iclr2027_conference,times}

\usepackage{amsmath,amsfonts,bm}

\def\eqref#1{equation~\ref{#1}}

\def\1{\bm{1}}

\DeclareMathAlphabet{\mathsfit}{\encodingdefault}{\sfdefault}{m}{sl}
\SetMathAlphabet{\mathsfit}{bold}{\encodingdefault}{\sfdefault}{bx}{n}

\usepackage{graphicx}
\usepackage{float}
\usepackage{colortbl}
\usepackage{pifont}
\usepackage{hyperref}
\usepackage{url}
\usepackage{etoolbox}

\makeatletter
\patchcmd{\@maketitle}
  {\hsize\textwidth}
  {\hsize\textwidth\centering}
  {}
  {\errmessage{Title centering patch failed}}
\patchcmd{\@maketitle}
  {\begin{tabular}[t]{l}\bf\rule{\z@}{24pt}\@author\end{tabular}}
  {\begin{tabular}[t]{c}\bf\rule{\z@}{16pt}\@author\end{tabular}}
  {}
  {\errmessage{Author centering patch failed}}
\patchcmd{\@maketitle}
  {\vskip 0.3in minus 0.1in}
  {\vskip 6pt}
  {}
  {\errmessage{Author spacing patch failed}}
\makeatother

\hypersetup{
  hidelinks,
  pdftitle={FrameGRPO: Incentivizing Video Reasoning via Multi-Frame Rollouts and Temporal Credit},
  pdfauthor={Jiaxuan Sun, Biao Wang, Shuai Li}
}

\title{FrameGRPO: Incentivizing Video Reasoning via Multi-Frame Rollouts and Temporal Credit}

\author{Jiaxuan Sun$^{1,2,3}$ \quad Biao Wang$^{3*}$ \quad Shuai Li$^{1,2*}$ \\
\normalfont $^{1}$Shanghai Jiao Tong University \\
\normalfont $^{2}$Shanghai Innovation Institute \quad $^{3}$Tencent Hy \\
\normalfont \texttt{jiaxuansun04@gmail.com} \quad \texttt{shuaili8@sjtu.edu.cn}
}

\iclrfinalcopy
\begin{document}

\maketitle
\begin{NoHyper}
  {\renewcommand{\thefootnote}{\ensuremath{*}}%
   \footnotetext{Corresponding authors}}
\end{NoHyper}

\fancypagestyle{arxivfirst}{%
  \fancyhf{}%
  \fancyhead[L]{\raisebox{-2pt}[0pt][0pt]{\includegraphics[height=16pt]{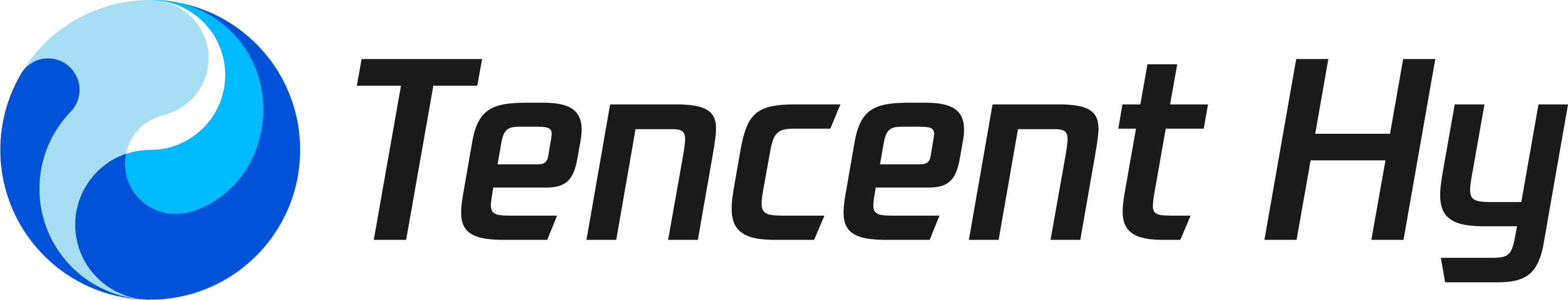}}}%
  \fancyfoot[C]{\thepage}%
  \renewcommand{\headrulewidth}{0.4pt}}
\thispagestyle{arxivfirst}
\lhead{}
\renewcommand{\headrulewidth}{0pt}

\begin{abstract}
Reinforcement learning with verifiable rewards, particularly Group Relative
Policy Optimization (GRPO), has improved video reasoning in multimodal large
language models. However, standard video GRPO varies only linguistic
trajectories while conditioning every rollout in a group on the same sampled
frames, which can produce homogeneous rewards and weaken group-relative
learning when the shared visual evidence is insufficient or distracting. We
propose \textbf{FrameGRPO}, a simple and effective reinforcement-learning
framework that requires neither auxiliary networks nor changes to the model
architecture. Specifically, FrameGRPO combines two lightweight components: (1)
\textbf{Multi-Frame Rollouts}, which use only uniform temporal subsampling of a
single maximal-frame decode to construct aligned views that diversify visual
evidence and reasoning trajectories; and (2) \textbf{Gated Temporal Efficiency
Credit}, which selectively reinforces a correct lower-ratio trajectory when all
rollouts in a fixed full-ratio reference subset fail, without imposing an
unconditional preference for fewer frames.
Extensive experiments
across diverse video understanding and reasoning benchmarks demonstrate robust
overall improvements over standard GRPO, highlighting temporal-view
diversification as an effective mechanism for exploration and credit
assignment in video MLLMs.
\end{abstract}

\section{Introduction}
\label{sec:introduction}

Recent progress in reinforcement learning with verifiable rewards (RLVR) has
advanced model reasoning beyond the imitation of curated solution traces toward
learning directly from outcome-level feedback
\citep{shao2024deepseekmath,guo2025deepseek}. Group Relative Policy
Optimization (GRPO) has become a particularly practical recipe because it learns
from relative comparisons among responses to the same query without requiring a
separate critic. This paradigm has rapidly expanded from language reasoning to
multimodal and video models \citep{feng2026video,wang2026incentivizing,fu2026videokr}.
Recent video-reasoning systems further improve fine-grained visual--temporal
credit assignment or allow models to acquire relevant evidence during inference
\citep{wang2026video,zhang2026thinking,xie2025video,ding2026videozoomer}.
Together, these advances establish RL post-training as a promising route from
passive video recognition toward reasoning grounded in temporal evidence.

Despite this progress, standard video GRPO explores variation primarily in the
language space while keeping the visual context fixed. Rollouts for the same
video--question pair use the same frame sampling ratio and therefore the same
temporal view, so stochastic decoding changes the linguistic trajectory but not
the density of visual input. Yet the frame sampling ratio governs an important
trade-off within the same decoded video: lower-ratio views contain fewer
redundant or irrelevant frames and can reduce visual distraction, whereas
higher-ratio views preserve more task-relevant temporal evidence. Fixing a
single frame sampling ratio forces every rollout to inherit the same balance
between evidence sufficiency and visual interference, making their outcomes
more correlated. When a group becomes uniformly correct or incorrect, GRPO has
little basis for relative comparison because all responses receive homogeneous
rewards, causing the task-relevant learning signal to vanish. This
\emph{advantage collapse} wastes rollouts and can stall optimization
\citep{he2026advantage, yu2026dapo, zhang2026scaf}. Existing difficulty-adaptive
methods restore variation by modifying or regrouping questions
\citep{gao2026diva}, but they do not address the shared visual context that is
distinctive to video reasoning.

\begin{figure}[t]
    \centering
    \includegraphics[width=\linewidth]{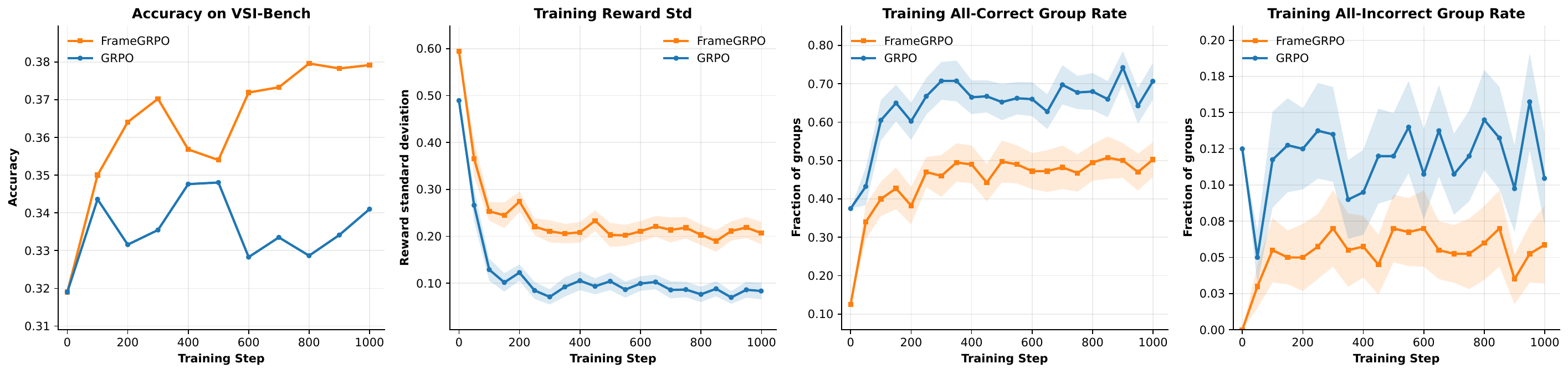}
    \caption{\textbf{Training dynamics of FrameGRPO
    and GRPO.} The four panels track VSI-Bench accuracy,
    within-group reward standard deviation, and the prevalence of uniformly
    correct and uniformly incorrect rollout groups during optimization. Exposing
    each group to multiple frame sampling ratios yields more informative reward
    contrasts, mitigates group-level saturation at either outcome extreme, and
    translates into stronger evaluation performance. Shaded bands visualize the
    variability of each metric.}
    \label{fig:training-dynamics}
\end{figure}

We propose \textbf{FrameGRPO} to address this evidence-induced advantage
collapse. Our key observation is that the frame sampling ratio is not merely a
computational setting: for the same video and question, it changes which events
are visible, how much distracting content is present, and which reasoning paths
are elicited. FrameGRPO therefore turns the frame sampling ratio into a
within-group exploration variable, allowing the policy to compare trajectories
grounded in different but temporally aligned views. This controlled evidence
diversity makes mixed outcomes more likely and preserves the relative signal
that GRPO requires. Figure~\ref{fig:training-dynamics} illustrates this
mechanism: compared with fixed-ratio GRPO, FrameGRPO maintains greater
within-group reward variation, substantially reduces both all-correct and
all-incorrect groups, and improves performance across video-reasoning
benchmarks.

FrameGRPO consists of two complementary modules. The \textbf{Multi-Frame
Rollouts} module first decodes each video once at the maximum sampled frame count,
assigns the resulting full-ratio view to a fixed subset of rollout slots, and
uniformly subsamples it for the remaining slots at randomly sampled frame
sampling ratios. The same policy generates responses from these views, and a
shared verifier evaluates them within one group, introducing diversity in both
visual evidence and reasoning trajectories. \textbf{Gated Temporal Efficiency
Credit} uses the outcomes of the fixed full-ratio subset to control credit
assignment. If any full-ratio rollout is correct, it leaves the standard GRPO
signal unchanged. Only when all full-ratio rollouts fail does it assign
additional credit to successful lower-ratio trajectories, with greater emphasis
on successes obtained at lower frame sampling ratios. Finally, every sampled
completion is conditioned on the full-ratio input for policy and reference
likelihood evaluation, transferring discoveries from complementary views into a
single policy. Together, the two modules create useful cross-view comparisons
without unconditionally favoring lower frame sampling ratios.

To validate the effectiveness of FrameGRPO, we evaluate it on seven diverse
video understanding and reasoning benchmarks: Video-MME \citep{fu2025video},
TempCompass \citep{liu2024tempcompass}, VideoEval-Pro \citep{ma2025videoeval},
MMVU \citep{zhao2025mmvu}, VSI-Bench \citep{yang2025thinking}, NExT-GQA
\citep{xiao2024can}, and Video-MMMU \citep{hu2026video}. The results show
that FrameGRPO outperforms standard GRPO across all evaluated settings,
demonstrating robust improvements across complementary aspects of video
reasoning.

Our contributions are threefold:
\begin{itemize}
    \item We propose \textbf{FrameGRPO}, a simple and effective reinforcement-learning
    framework that improves video reasoning.
    \item FrameGRPO explores multiple frame sampling ratios to diversify temporal
    evidence and introduces a cross-view credit assignment strategy that
    reinforces successful trajectories discovered from complementary
    ratio-conditioned views.
    \item Extensive experiments across seven video understanding and reasoning
    benchmarks demonstrate that FrameGRPO consistently surpasses standard GRPO,
    validating its effectiveness and robustness across diverse video 
    tasks.
\end{itemize}

\section{Related Work}
\label{sec:related-work}

\subsection{MLLMs for Video Understanding}

Video MLLMs must distill long, redundant streams into tractable evidence:
sparse sampling can miss decisive events, whereas dense inputs increase cost
and distraction \citep{zhu2026focus,zou2026videobrain,he2026framethinker}.
Existing work addresses this trade-off through compact visual memory
\citep{chen2026visualmemory}, retrieval
\citep{luo2025videorag,zhang2025deepvideo,bahrami2026strive}, keyframe or
adaptive sampling \citep{li2026selectless,qin2026frameselection,
qiu2026longvideor1}, and token compression
\citep{xu2026streamingvlm,wu2026marc}.  These strategies reduce redundancy
before or during decoding but primarily optimize the evidence available to each
individual reasoning trajectory.  Complementary methods integrate evidence
acquisition into reasoning through frame-wise chains of thought
\citep{arnab2025temporal}, iterative search or temporal zooming
\citep{he2026framethinker,pan2026timesearch,ding2026videozoomer}, and agentic
tool use \citep{zeng2026videoo3,xie2025video,zhang2026thinking,jain2026sage}.
Temporal objectives, grounding post-training, and evidence-level rewards further
improve temporal supervision \citep{wu2026tempr1,wang2025timer1,li2026videoopd,
meng2026openo3video,cao2026videotrace}.  Unlike these approaches, FrameGRPO
uses aligned multi-ratio views within each GRPO group as an informative signal
for a single policy, requiring neither auxiliary modules nor additional
frame-decoding computation.

\subsection{Reinforcement Learning in MLLMs}

RLVR methods such as GRPO optimize relative outcomes within groups without a
learned critic \citep{shao2024deepseekmath,guo2025deepseek,yu2026dapo}.  In
MLLMs, they support general or grounded visual reasoning
\citep{yu2025perceptionr1,liu2026visionreasoner,sarch2025grounded,
yao2025r1sharevl}, active perception and tool use
\citep{wu2026vtoolr1,chng2026sensesearch,zhu2026activeo3}, and alternative
optimization signals \citep{xiao2025fastslow,yoon2026pdcr,du2026appo}.  Video
RL additionally requires deciding which temporal evidence to inspect.  Prior
systems establish reinforced fine-tuning
\citep{feng2026video,wang2025videorft,chen2025scalingrl}, improve temporal
grounding and sampling \citep{li2025tempsampr1,wang2025timer1,
qin2026frameselection}, enable active navigation
\citep{qiu2026longvideor1,he2026framethinker,pan2026timesearch}, or provide
evidence-aware supervision \citep{wang2026video,cao2026videotrace,
linghu2026threedrft}.  Nevertheless, when all rollouts share a fixed video view,
insufficient or distracting evidence can still yield uniform rewards and erase
the relative learning signal.  Unlike prior remedies for uninformative
uniform-reward groups
\citep{yu2026dapo,park2025deepvideor1,gao2026diva,he2026advantage}, FrameGRPO
targets the video-specific cause---identical visual context across
rollouts---using multiple frame-sampling views and Gated Temporal Efficiency
Credit.

\section{FrameGRPO}
\label{sec:framegrpo}

\begin{figure}[t]
    \centering
    \includegraphics[width=\linewidth]{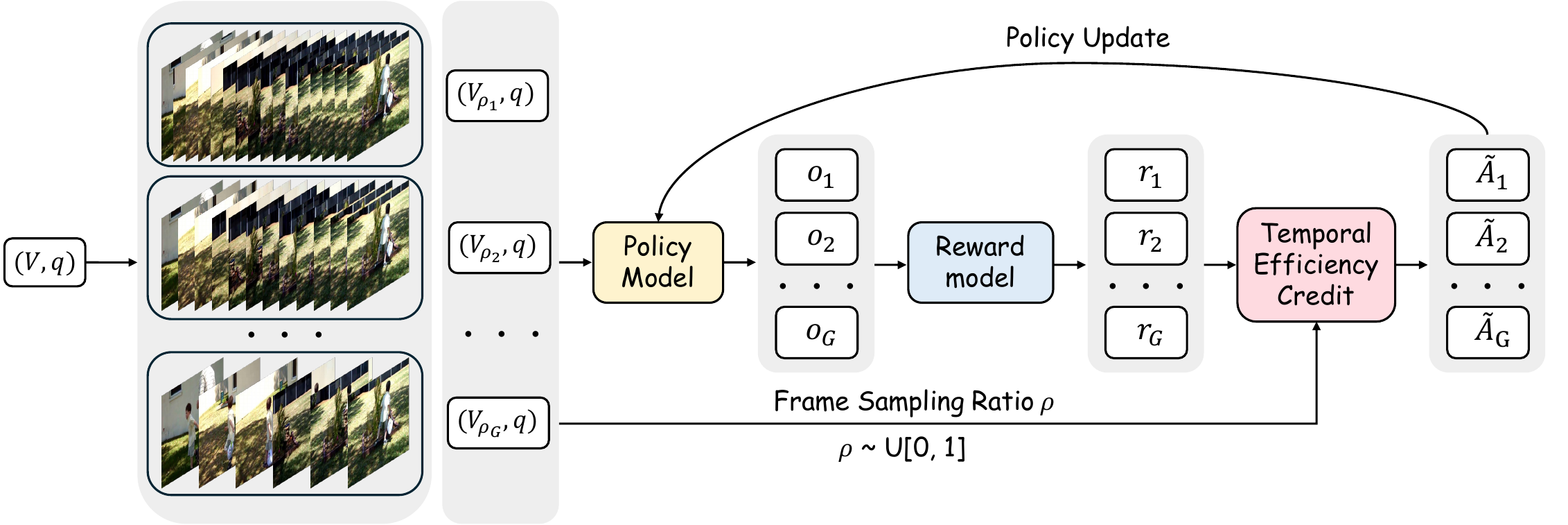}
    \caption{Overview of FrameGRPO. Given a video--question pair, we assign the
    full-ratio input to a fixed subset of rollout slots and independently sample
    $\rho_i\sim\mathcal{U}[0,1]$ for the remaining slots to construct temporally
    aligned views $V_{\rho_i}$. A shared policy generates
    $\{o_i\}_{i=1}^{G}$, and a verifier assigns rewards $\{r_i\}_{i=1}^{G}$.
    The outcomes of the fixed full-ratio subset jointly determine the gate for
    Temporal Efficiency Credit, which produces the modified advantages
    $\{\widetilde A_i\}_{i=1}^{G}$. Policy optimization is conditioned only on
    the full-ratio input.}
    \label{fig:framegrpo-overview}
\end{figure}

In this section, we introduce \textbf{FrameGRPO}, a simple and effective
reinforcement-learning framework for improving video reasoning in multimodal
large language models (MLLMs). To address the limited visual exploration of
vanilla GRPO and the resulting risk of advantage collapse, FrameGRPO explores
multiple frame sampling ratios and selectively reinforces successful
trajectories from lower-ratio visual contexts via gated temporal efficiency
credit when all rollouts in a fixed full-ratio reference subset fail.
Section~\ref{sec:grpo-preliminaries} reviews GRPO and its evidence-induced
advantage collapse; Section~\ref{sec:multi-frame} describes multi-frame rollout
construction; and Section~\ref{sec:temporal-credit} introduces our gated credit
mechanism. Figure~\ref{fig:framegrpo-overview} illustrates the overall pipeline.

\subsection{Preliminaries: Group Relative Policy Optimization}
\label{sec:grpo-preliminaries}

Let $V$ denote a sampled video view, $q$ a question, and $a$ its reference
answer. Given the fixed multimodal prompt $(V,q)$, vanilla video GRPO samples a
group of $G$ responses from the old policy,
$o_i\sim\pi_{\theta_{\mathrm{old}}}(\cdot\mid V,q)$ for
$i\in\{1,\ldots,G\}$. A shared verifier assigns each response a scalar reward
$r_i=R(o_i,a)$. In our instantiation, $r_i$ is the sum of a correctness reward
and a format reward; we retain the correctness signal separately for the gated
credit introduced in Section~\ref{sec:temporal-credit}. Let
$\mathbf{r}=(r_1,\ldots,r_G)$ collect the rewards of all responses. GRPO avoids
a learned critic and estimates the normalized advantage as
\begin{equation}
    \widehat A_i
    = \frac{r_i-\operatorname{mean}(\mathbf{r})}
    {\operatorname{std}(\mathbf{r})+\delta},
    \label{eq:grpo-advantage}
\end{equation}
where $\delta>0$ is a small constant for numerical stability. The
KL-regularized GRPO objective with a PPO-style clipped update is
\begin{equation}
\begin{aligned}
    \mathcal{J}_{\mathrm{GRPO}}(\theta)
    ={}& \mathbb{E}_{\substack{(V,q,a)\sim\mathcal{D},\\
        \{o_i\}_{i=1}^{G}\sim\pi_{\theta_{\mathrm{old}}}(\cdot\mid V,q)}}
    \Bigg[\frac{1}{G}\sum_{i=1}^{G}\Bigg(
    \min\Bigg\{
    \frac{\pi_\theta(o_i\mid V,q)}
    {\pi_{\theta_{\mathrm{old}}}(o_i\mid V,q)}\widehat A_i,\\[-0.2em]
    &\qquad \operatorname{clip}\!\left(
    \frac{\pi_\theta(o_i\mid V,q)}
    {\pi_{\theta_{\mathrm{old}}}(o_i\mid V,q)},
    1-\varepsilon,1+\varepsilon
    \right)\widehat A_i\Bigg\}
    -\beta\widehat D_{\mathrm{KL}}^{(i)}
    (\pi_\theta\,\|\,\pi_{\mathrm{ref}})
    \Bigg)\Bigg],
    \label{eq:grpo-objective}
\end{aligned}
\end{equation}
where $\pi_\theta$ is the current policy, $\varepsilon>0$ sets the clipping
range, $\pi_{\mathrm{ref}}$ is a frozen reference policy, and $\beta$ controls
the KL penalty.

Equation~\ref{eq:grpo-advantage} relies on heterogeneous rewards. If every
response is assigned the same reward, then both the centered rewards and their
standard deviation vanish, yielding $\widehat A_i=0$ and no
reward-driven gradient. This failure mode is commonly referred to as
\emph{advantage collapse} \citep{he2026advantage, zhang2026scaf}. It is particularly relevant
to video GRPO: every rollout observes exactly the same frames, so stochastic
decoding explores only the language trajectory while leaving the underlying
visual evidence unchanged. An insufficient or distracting view can therefore
make the entire group fail in a correlated manner. FrameGRPO addresses this
modality-specific degeneracy by making the frame sampling ratio an additional
within-group exploration variable.

\subsection{Multi-Frame Rollouts for Temporal Exploration}
\label{sec:multi-frame}

\begin{figure}[t]
    \centering
    \includegraphics[width=\linewidth]{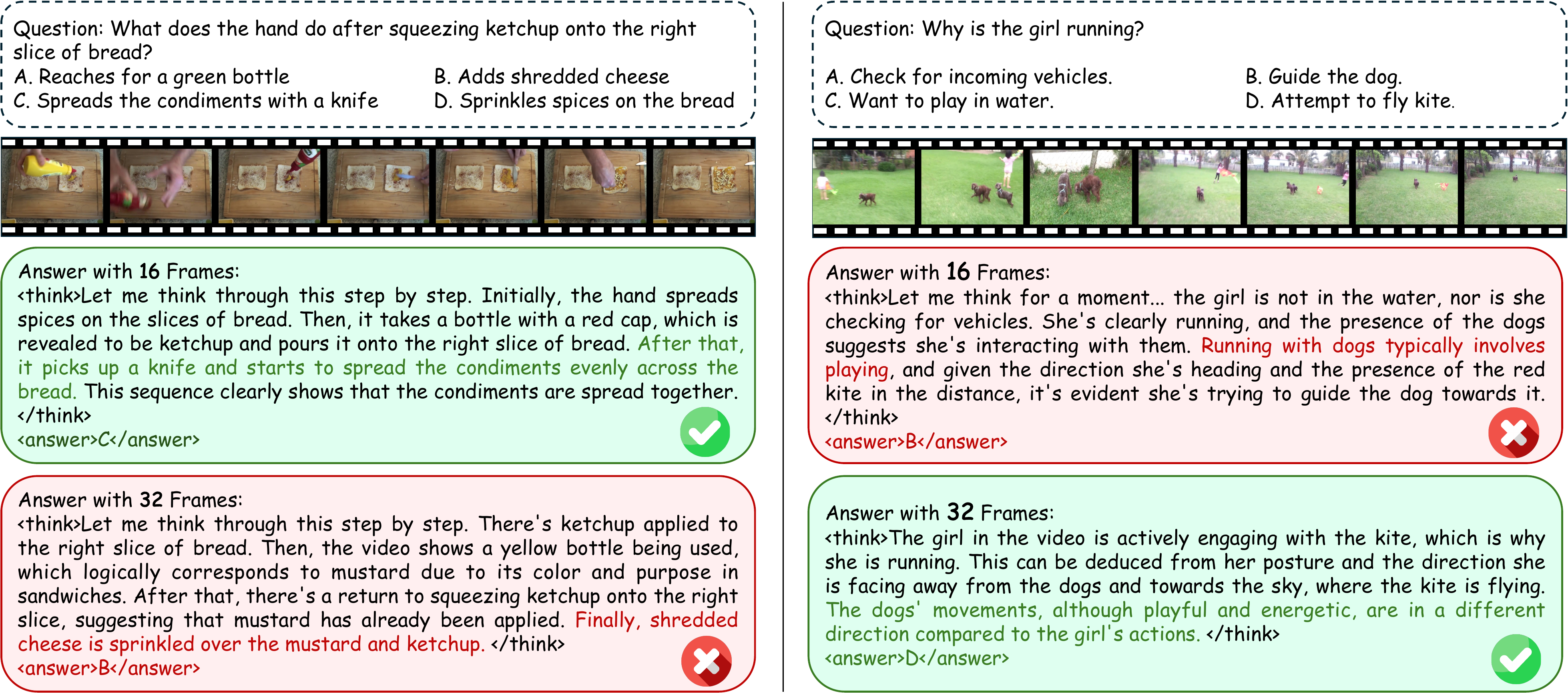}
    \caption{Complementarity across frame sampling ratios. \emph{Left:} the
    lower-ratio 16-frame view captures the relevant condiment-spreading action,
    whereas the full-ratio 32-frame view is distracted by a subsequent event.
    \emph{Right:} the higher-ratio view preserves the evidence required to
    associate the girl's motion with the kite, whereas the lower-ratio view
    misses this critical evidence.}
    \label{fig:frame-complementarity}
\end{figure}

\paragraph{Complementary temporal evidence.}
The frame sampling ratio controls which temporal evidence enters the model, not
merely its computational cost. Because relevant and distracting events are
unevenly distributed over time, changing this ratio can affect reasoning in a
non-monotonic manner, as illustrated in Figure~\ref{fig:frame-complementarity}.
We therefore vary the ratio across rollouts in each group and use the resulting
complementary temporal views as a source of exploration.

\paragraph{Multi-frame rollout strategy.}
Let $f_\star$ denote the maximum number of sampled frames and $V_\rho$ the view
associated with a frame sampling ratio $\rho\in[0,1]$; the full-ratio view $V_1$
therefore contains $f_\star$ frames. To explore temporal evidence while
retaining a reliable reference for credit assignment, every multi-frame group
reserves a fixed subset of rollout slots for $V_1$. Let
$\mathcal{I}_\star\subset\{1,\ldots,G\}$ index this subset and
$K_\star=|\mathcal{I}_\star|$, where $1<K_\star<G$. We set
$\rho_i=1$ for $i\in\mathcal{I}_\star$ and independently sample
$\rho_i\sim\mathcal{U}[0,1]$ for the remaining $G-K_\star$ slots. Each sampled
ratio is mapped to an integer frame count $f_i=\lceil\rho_i f_\star\rceil$; after
rounding, we use the realized ratio $\rho_i\leftarrow f_i/f_\star$ thereafter.
We decode the video once at $f_\star$ frames and uniformly subsample $f_i$ frames
to construct $V_{\rho_i}$. All views cover the same temporal interval and
preserve temporal order. The full-ratio slots share the
complete sampled context $V_1$ but yield multiple stochastic rollouts, whereas
the remaining slots provide ratio-conditioned exploration.

Conditioned on each view $V_{\rho_i}$, the old policy samples a rollout
$o_i\sim\pi_{\theta_{\mathrm{old}}}(\cdot\mid V_{\rho_i},q)$, which receives the
verifier reward $r_i=R(o_i,a)$. Together, the full-ratio and exploratory subsets
produce $G$ responses $\{o_i\}_{i=1}^{G}$ grounded in complementary temporal
contexts; the associated ratios $\{\rho_i\}_{i=1}^{G}$ and rewards
$\{r_i\}_{i=1}^{G}$ are retained for subsequent credit assignment. In
particular, the outcomes $\{(o_i,r_i)\}_{i\in\mathcal{I}_\star}$ provide the
full-ratio reference required by Gated Temporal Efficiency Credit.

Ratio-conditioned views are used only during rollout collection; policy
optimization is conditioned exclusively on the full-ratio input $(V_1,q)$ to
ensure stable learning and preserve the accuracy of the learned policy.

\subsection{Gated Temporal Efficiency Credit}
\label{sec:temporal-credit}

Multi-frame rollouts increase outcome diversity, but standard group
normalization does not explicitly distinguish between a correct trajectory
discovered from a lower-ratio view and those obtained from the fixed full-ratio
reference subset. This subset makes the distinction observable in every group:
when all of its rollouts fail but a lower-ratio view succeeds, the latter reveals
a useful trajectory absent from the sampled full-ratio outcomes. We introduce
\textbf{Gated Temporal Efficiency Credit} (TEC) to retain such discoveries
without imposing an unconditional preference for lower frame sampling ratios.

We distinguish the scalar verifier reward $r_i$, which may combine correctness
and format, from a binary correctness indicator
$c_i=\mathbf{1}[o_i\text{ answers }q\text{ correctly}]$. The gate activates only
when every rollout in the fixed full-ratio subset is incorrect:
\begin{equation}
    g=\mathbf{1}\!\left[
    \max_{j\in\mathcal{I}_\star}c_j=0\right].
    \label{eq:tec-gate}
\end{equation}
For each trajectory, TEC assigns the bonus
\begin{equation}
    b_i=\alpha\,g\,c_i
    \log\!\left(\frac{1}{\rho_i}\right),
    \qquad
    \widetilde A_i=\widehat A_i+b_i,
    \label{eq:temporal-credit}
\end{equation}
where $\widehat A_i$ is the standard group-relative advantage in
Equation~\ref{eq:grpo-advantage} and $\alpha\geq0$ controls the credit strength.
We use $\alpha=0.2$ by default. The inverse frame sampling ratio gives larger
credit to a correct trajectory discovered from a lower-ratio view and is zero
when $\rho_i=1$.

We next define the complete FrameGRPO objective. Let
$p_{\mathrm{MF}}(\boldsymbol{\rho})$ denote the group-level ratio assignment
described above: $\rho_i=1$ for $i\in\mathcal{I}_\star$, while the remaining
ratios are independently sampled. The objective is
\begin{equation}
\begin{aligned}
    \mathcal{J}_{\mathrm{FrameGRPO}}(\theta)
    ={}& \mathbb{E}_{\substack{(V,q,a)\sim\mathcal{D},\,
        \boldsymbol{\rho}\sim p_{\mathrm{MF}}\\
        o_i\sim\pi_{\theta_{\mathrm{old}}}
        (\cdot\mid V_{\rho_i},q),\ i=1,\ldots,G}}
    \Bigg[\frac{1}{G}\sum_{i=1}^{G}\Bigg(
    \min\Bigg\{
    \frac{\pi_\theta(o_i\mid V_1,q)}
    {\pi_{\theta_{\mathrm{old}}}(o_i\mid V_1,q)}\widetilde A_i,\\[-0.2em]
    &\qquad \operatorname{clip}\!\left(
    \frac{\pi_\theta(o_i\mid V_1,q)}
    {\pi_{\theta_{\mathrm{old}}}(o_i\mid V_1,q)},
    1-\varepsilon,1+\varepsilon
    \right)\widetilde A_i\Bigg\}
    -\beta\widehat D_{\mathrm{KL}}^{(i)}
    (\pi_\theta\,\|\,\pi_{\mathrm{ref}})
    \Bigg)\Bigg].
    \label{eq:framegrpo-objective}
\end{aligned}
\end{equation}

The gate yields three desirable behaviors. If any rollout in the full-ratio
reference subset succeeds, then $g=0$ and
$\widetilde A_i=\widehat A_i$, leaving the standard GRPO advantage unchanged.
If all full-ratio reference rollouts fail but a lower-ratio view succeeds, only
the correct lower-ratio trajectories receive additional credit. If every view
fails, $c_i=0$ for all trajectories and no bonus is introduced. Thus, TEC
exploits a lower-ratio success only when it reveals a trajectory not discovered
by any rollout in the fixed full-ratio subset; it does not reward a lower frame
sampling ratio by default. Since the mechanism changes only the training-time
advantages, it introduces no additional inference cost.

\section{Experiments}
\label{sec:experiments}

\begin{table*}
\caption{\textbf{Performance Comparison on Video Understanding and Reasoning Benchmarks.} All results are reported as accuracy (\%). The best result within each backbone is highlighted in bold.}
  \label{tab:video_benchmarks}
  \centering
  \normalsize
  \setlength{\tabcolsep}{1.5pt}
  \renewcommand{\arraystretch}{1.2}
\resizebox{\textwidth}{!}{%
\begin{tabular}{lccc|cccc}
\hline
\raisebox{-5.5pt}[0pt][0pt]{{\small\textbf{Method}}} & \multicolumn{3}{c|}{{\small\textbf{Video General Benchmark}}} & \multicolumn{4}{c}{{\small\textbf{Video Reasoning Benchmark}}} \\ \cline{2-8}
 & {\small\textbf{Video-MME}} & {\small\textbf{TempCompass}} & {\small\textbf{VideoEval-Pro}} & {\small\textbf{MMVU}} & {\small\textbf{VSI-Bench}} & {\small\textbf{NExT-GQA}} & {\small\textbf{Video-MMMU}} \\ \hline
GPT-4o                & 71.9 & 75.1 & 34.2 & 75.4 & 34.0 & -    & 61.2 \\
Gemini 1.5 Pro        & 75.0 & 70.5 & 39.3 & 71.2 & 48.8 & -    & 53.4 \\
LLaMA-VID             & -    & 45.6 & -    & -    & -    & -    & -    \\
LLaVA-OneVision-7B    & 58.2 & 69.5 & -    & 49.2 & 32.4 & -    & 33.8 \\
Kangaroo-8B           & 56.0 & 62.5 & -    & -    & -    & -    & -    \\
VILA-1.5-8B           & -    & 58.8 & -    & -    & 28.9 & -    & 20.8 \\
Video-R1-7B           & 60.0 & 73.2 & 42.6 & 63.4 & 36.7 & 77.3 & 51.8 \\
VideoRFT              & 60.7 & 73.6 & 40.8 & 66.1 & 37.2 & 78.0 & 49.6 \\
VideoChat-R1          & 59.6 & 73.8 & 39.3 & 64.3 & 36.3 & 78.3 & 51.3 \\
Time-R1-7B            & 59.6 & 73.3 & 40.3 & 63.5 & 24.9 & 78.6 & 51.9 \\ \hline
\rowcolor{yellow!10}\multicolumn{8}{c}{\textbf{Main Results}} \\ \hline
Qwen2.5-VL-7B-Instruct & 57.4 & 72.9 & 38.6 & 61.4 & 31.9 & 76.9 & 51.7 \\
+ SFT                   & 53.8 & 69.5 & 36.5 & 61.6 & 31.9 & 70.0 & 43.6 \\
+ GRPO                  & 60.0 & 74.4 & 41.0 & 65.6 & 34.1 & 79.5 & 52.0 \\
\rowcolor{blue!6}+ FrameGRPO   & \textbf{61.3} & \textbf{74.9} & \textbf{42.4} & \textbf{66.7} & \textbf{37.9} & \textbf{79.6} & \textbf{54.2} \\ \hline
Qwen2-VL-7B-Instruct   & 57.4 & 70.1 & 42.0 & 59.2 & 23.2 & 77.2 & 31.3 \\
+ SFT                  & 52.3 & 67.4 & 39.1 & 56.8 & \textbf{29.1} & 67.7 & 35.8 \\
+ GRPO                 & 57.9 & 71.3 & 43.2 & 61.6 & 27.6 & 76.9 & 41.3 \\
\rowcolor{blue!6}+ FrameGRPO  & \textbf{58.4} & \textbf{72.3} & \textbf{44.8} & \textbf{62.2} & 27.4 & \textbf{77.3} & \textbf{43.6} \\ \hline
\end{tabular}%
}
\vspace{-4mm}
\end{table*}

\subsection{Training Data}
\label{sec:training-data}
We construct our training set exclusively from the video portion of
Video-R1-260K \citep{feng2026video}, which contains approximately 116K video
samples collected from diverse open-domain scenarios. These data cover event
progression, object motion, temporal dependencies, action understanding, and
causal reasoning, providing broad supervision for spatiotemporal video
understanding. Following prior work \citep{yu2026dapo,zheng2026deepeyes}, we
perform difficulty-based data curation by generating eight responses for each candidate
question and estimating its difficulty from the empirical accuracy. We exclude
questions with an accuracy of either zero or one, corresponding to cases where
none or all of the responses are correct. After filtering, we obtain 8K
examples as our final training set.

\subsection{Implementation Details}
\label{sec:implementation-details}
We implement FrameGRPO on top of the Open-R1-Video \citep{wang2026incentivizing} reinforcement learning
framework and conduct experiments with Qwen2.5-VL-7B and Qwen2-VL-7B as the
base MLLMs, comparing standard GRPO against our proposed FrameGRPO for each
backbone. For training efficiency, we cap the number of video frames at 32 and
treat the uniformly sampled 32-frame sequence as the full-ratio view. Each
FrameGRPO group assigns this fixed input to a subset $\mathcal{I}_\star$ of
rollout slots; for every remaining slot, we independently draw
$\rho_i\sim\mathcal{U}[0,1]$ and uniformly subsample the full-ratio sequence to
construct the corresponding temporally aligned view $V_{\rho_i}$. Each frame is
processed with a maximum pixel budget of
$256\times28\times28$. Each training run takes approximately 21 hours on a
single node equipped with eight NVIDIA H20 GPUs.

\subsection{Evaluation Benchmarks}
\label{sec:evaluation-benchmarks}
We evaluate our models on seven benchmarks covering general video understanding
and video reasoning. The video general benchmarks include: (1) Video-MME
\citep{fu2025video}, with 2.7K questions over 900 videos ranging from 11 seconds
to one hour across diverse domains; (2) TempCompass \citep{liu2024tempcompass},
with 7,540 instructions over 410 videos assessing action, direction, speed,
event order, and attribute change; and (3) VideoEval-Pro
\citep{ma2025videoeval}, with 1,289 open-ended questions over 465 videos longer
than ten minutes for realistic long-video understanding. The video reasoning
benchmarks include: (4) MMVU \citep{zhao2025mmvu}, with 3K expert-level
questions across 27 disciplines requiring visual and domain-knowledge
reasoning; (5) VSI-Bench \citep{yang2025thinking}, with over 5K questions from
288 egocentric indoor videos for spatial and spatiotemporal reasoning; (6)
NExT-GQA \citep{xiao2024can}, with 8,911 causal and temporal questions paired
with grounded temporal evidence; and (7) Video-MMMU \citep{hu2026video}, with
900 questions over 300 college-level educational videos assessing perception,
comprehension, and knowledge transfer across six disciplines.

\subsection{Baselines}
\label{sec:baselines}
To provide a broad comparison, we report results from proprietary models,
including GPT-4o \citep{openai2024gpt4o} and Gemini 1.5 Pro
\citep{geminiteam2024gemini}, as well as general-purpose open-source video MLLMs,
including LLaMA-VID \citep{li2023llamavid}, LLaVA-OneVision-7B
\citep{li2024llavaonevision}, Kangaroo-8B \citep{liu2024kangaroo}, and
VILA-1.5-8B \citep{lin2024vila}.
We further compare with recent reinforcement-learning-based video reasoning
models, including Video-R1-7B \citep{feng2026video}, VideoRFT
\citep{wang2025videorft}, VideoChat-R1 \citep{wang2026incentivizing}, and
Time-R1-7B \citep{wang2025timer1}. For controlled comparisons on both
Qwen2.5-VL-7B and Qwen2-VL-7B backbones, we include the original instruction
model, supervised fine-tuning on the same training set, and standard GRPO under
the same experimental configuration, allowing us to isolate the contribution
of the proposed frame-aware exploration and credit assignment.

\subsection{Performance Analysis}
\label{sec:performance-analysis}
Table~\ref{tab:video_benchmarks} compares FrameGRPO with the baseline methods
across general video understanding and video reasoning benchmarks. On the
general benchmarks, FrameGRPO consistently improves upon standard GRPO for both
backbones, with gains spanning comprehensive video question answering,
fine-grained temporal understanding, and open-ended long-video evaluation. This
shows that varying the frame sampling ratio during rollout collection improves
temporal exploration without sacrificing the broad video-understanding ability
of the underlying model.

FrameGRPO also exhibits strong performance on the reasoning benchmarks, which
cover expert-level multidisciplinary reasoning, spatial and spatiotemporal
understanding, temporally grounded question answering, and knowledge acquisition
from educational videos. It surpasses GRPO on every benchmark with
Qwen2.5-VL-7B and in nearly all settings with Qwen2-VL-7B, while achieving
leading performance among similar-scale open-source methods on most benchmarks.
The consistent trend across two model families suggests that the improvements
are not specific to a single backbone. Moreover, the broad reasoning gains
support our motivation that complementary temporal views can expose useful
trajectories that are difficult to discover when all rollouts share a fixed
visual context, while gated credit assignment effectively incorporates these
discoveries into policy learning.

We further observe that supervised fine-tuning on the same training set does not
provide consistent gains and often degrades the base models. In contrast,
standard GRPO yields substantial improvements, which FrameGRPO further
strengthens across both general understanding and reasoning tasks. This
comparison indicates that the performance gains arise primarily from
reinforcement learning and the proposed frame-aware exploration and credit
assignment, rather than from exposure to the training data alone.

\subsection{Ablation Study}
\label{sec:ablation-study}

\paragraph{Method Design.}
To examine the contribution of each component in FrameGRPO, we compare standard
GRPO, GRPO augmented only with multi-frame rollouts, and the complete method
combining multi-frame rollouts with Gated Temporal Efficiency Credit. As shown
in Table~\ref{tab:module_ablation}, introducing multi-frame rollouts consistently
improves performance across the evaluation suite, confirming that varying the
visual context within a rollout group broadens temporal exploration and provides
more informative group-relative supervision. Adding Gated Temporal Efficiency
Credit yields further improvements on most tasks and produces the strongest
overall performance. The consistent gains on most benchmarks demonstrate that
the two components are complementary: multi-frame rollouts expose useful
reasoning trajectories from alternative temporal views, while the credit
mechanism selectively preserves and reinforces these discoveries when all
rollouts in the fixed full-ratio reference subset fail.

\begin{table}
\caption{\textbf{Ablation Study of FrameGRPO Modules.} MFRoll and TempCred denote Multi-Frame Rollouts and Temporal Credit, respectively. All results are reported as accuracy (\%).}
\label{tab:module_ablation}
\centering
\small
\setlength{\tabcolsep}{1.4pt}
\renewcommand{\arraystretch}{1.2}
\resizebox{\textwidth}{!}{%
\begin{tabular}{cc|ccc|cccc}
\hline
\multicolumn{2}{c|}{{\footnotesize\textbf{Modules}}} & \multicolumn{3}{c|}{{\footnotesize\textbf{Video General Benchmark}}} & \multicolumn{4}{c}{{\footnotesize\textbf{Video Reasoning Benchmark}}} \\ \cline{1-9}
{\footnotesize\textbf{MFRoll}} & {\footnotesize\textbf{TempCred}} & {\footnotesize\textbf{Video-MME}} & {\footnotesize\textbf{TempCompass}} & {\footnotesize\textbf{VideoEval-Pro}} & {\footnotesize\textbf{MMVU}} & {\footnotesize\textbf{VSI-Bench}} & {\footnotesize\textbf{NExT-GQA}} & {\footnotesize\textbf{Video-MMMU}} \\ \hline
\ding{55} & \ding{55} & 60.0 & 74.4 & 41.0 & 65.6 & 34.1 & 79.5 & 52.0 \\
\ding{51} & \ding{55} & 60.7 & 74.4 & 41.8 & 66.3 & 34.7 & \textbf{79.8} & 52.2 \\
\rowcolor{blue!6}\ding{51} & \ding{51} & \textbf{61.3} & \textbf{74.9} & \textbf{42.4} & \textbf{66.7} & \textbf{37.9} & 79.6 & \textbf{54.2} \\ \hline
\end{tabular}%
}
\vspace{-3mm}
\end{table}

\paragraph{Policy Optimization Input.}
FrameGRPO uses ratio-conditioned views for rollout exploration but conditions
policy optimization only on the full-ratio input. To validate this design, we
compare it with a variant that computes the policy and reference likelihoods
using only the partial-ratio input associated with each rollout. As shown in
Table~\ref{tab:optimization_input_ablation}, conditioning optimization on the
full-ratio input matches or surpasses the partial-ratio alternative across all
benchmarks, with particularly clear advantages on spatial reasoning and
knowledge-acquisition tasks. These results indicate that partial-ratio views are
valuable for discovering complementary trajectories, whereas reattaching those
trajectories to the complete visual context during optimization more effectively
transfers the discovered reasoning behavior into the policy.

\begin{table}
\caption{\textbf{Ablation Study of Policy Optimization Input.} We compare conditioning policy optimization only on the partial-ratio input with conditioning only on the full-ratio input. All results are reported as accuracy (\%).}
\label{tab:optimization_input_ablation}
\centering
\small
\setlength{\tabcolsep}{1.4pt}
\renewcommand{\arraystretch}{1.2}
\resizebox{\textwidth}{!}{%
\begin{tabular}{l|ccc|cccc}
\hline
\raisebox{-5.5pt}[0pt][0pt]{{\footnotesize\textbf{Optimization Input}}} & \multicolumn{3}{c|}{{\footnotesize\textbf{Video General Benchmark}}} & \multicolumn{4}{c}{{\footnotesize\textbf{Video Reasoning Benchmark}}} \\ \cline{2-8}
 & {\footnotesize\textbf{Video-MME}} & {\footnotesize\textbf{TempCompass}} & {\footnotesize\textbf{VideoEval-Pro}} & {\footnotesize\textbf{MMVU}} & {\footnotesize\textbf{VSI-Bench}} & {\footnotesize\textbf{NExT-GQA}} & {\footnotesize\textbf{Video-MMMU}} \\ \hline
Partial-Ratio Input & 60.2 & 74.0 & 41.0 & \textbf{66.7} & 32.9 & 79.4 & 50.1 \\
\rowcolor{blue!6}Full-Ratio Input (Ours) & \textbf{61.3} & \textbf{74.9} & \textbf{42.4} & \textbf{66.7} & \textbf{37.9} & \textbf{79.6} & \textbf{54.2} \\ \hline
\end{tabular}%
}
\vspace{-3mm}
\end{table}

\paragraph{Hyperparameter Sensitivity.}
We investigate the sensitivity of FrameGRPO to $\alpha$, which controls the
strength of Gated Temporal Efficiency Credit. As shown in
Table~\ref{tab:alpha_sensitivity}, FrameGRPO retains improvements over standard
GRPO on most benchmarks across the tested range, demonstrating that its
effectiveness is robust to the choice of $\alpha$. A smaller credit weight
performs favorably on several general and expert-level reasoning tasks, whereas
the default moderate setting provides the strongest overall balance, with
particularly clear benefits on challenging spatial and knowledge-acquisition
tasks. Increasing the weight further weakens performance on some tasks, as an
excessive credit may overemphasize lower-ratio successes relative to the
standard group-relative learning signal. Overall, FrameGRPO does not require
precise tuning of $\alpha$, while a moderate value offers a reliable trade-off
across diverse video understanding and reasoning settings.

\begin{table}
\caption{\textbf{Sensitivity Analysis of Hyperparameter $\alpha$.} Results under varying $\alpha$ settings are reported on seven video benchmarks. The best performance for each benchmark is highlighted in bold. The configuration highlighted in blue corresponds to the setting used in the main results.}
\label{tab:alpha_sensitivity}
\centering
\small
\setlength{\tabcolsep}{1.4pt}
\renewcommand{\arraystretch}{1.2}
\resizebox{\textwidth}{!}{%
\begin{tabular}{l|ccc|cccc}
\hline
\raisebox{-5.5pt}[0pt][0pt]{{\footnotesize\textbf{Setting}}} & \multicolumn{3}{c|}{{\footnotesize\textbf{Video General Benchmark}}} & \multicolumn{4}{c}{{\footnotesize\textbf{Video Reasoning Benchmark}}} \\ \cline{2-8}
 & {\footnotesize\textbf{Video-MME}} & {\footnotesize\textbf{TempCompass}} & {\footnotesize\textbf{VideoEval-Pro}} & {\footnotesize\textbf{MMVU}} & {\footnotesize\textbf{VSI-Bench}} & {\footnotesize\textbf{NExT-GQA}} & {\footnotesize\textbf{Video-MMMU}} \\ \hline
Vanilla GRPO & 60.0 & 74.4 & 41.0 & 65.6 & 34.1 & 79.5 & 52.0 \\ \hline
\rowcolor{yellow!10}\multicolumn{8}{c}{\textbf{FrameGRPO with Different $\alpha$}} \\ \hline
$\alpha=0.1$ & \textbf{61.3} & \textbf{75.3} & \textbf{43.1} & \textbf{67.7} & 35.8 & \textbf{79.6} & 52.6 \\
\rowcolor{blue!6}$\alpha=0.2$ & \textbf{61.3} & 74.9 & 42.4 & 66.7 & \textbf{37.9} & \textbf{79.6} & \textbf{54.2} \\
$\alpha=0.3$ & 61.0 & 75.0 & 41.5 & 65.4 & 34.3 & \textbf{79.6} & 52.0 \\ \hline
\end{tabular}%
}
\vspace{-3mm}
\end{table}

\section{Limitation}
\label{sec:limitation}


FrameGRPO derives its additional training signal from differences among
temporally aligned views, and its benefit therefore depends on how strongly a
question responds to changes in frame density. For appearance-centric questions
that can be answered from a few isolated frames, varying the sampling ratio may
leave the rollout distribution nearly unchanged and provide little extra reward
diversity. At the other extreme, aggressive subsampling can omit brief but
decisive events when precise temporal localization is required. 
A promising direction is to estimate the
informativeness and consistency of each temporal view and adapt both view
construction and cross-view credit to individual video--question pairs.

\section{Conclusion}
\label{sec:conclusion}

In this work, we developed \textbf{FrameGRPO} to improve reinforcement learning
for video reasoning by extending policy exploration from linguistic trajectories
to temporal evidence. Its Multi-Frame Rollouts reuse a maximal-frame decode to
construct temporally aligned views at different sampling ratios, enabling more
informative cross-view comparisons within each GRPO group. To exploit useful
discoveries from sparse views without systematically favoring fewer frames,
Gated Temporal Efficiency Credit selectively strengthens a correct lower-ratio
trajectory only when all rollouts in the fixed full-ratio reference subset are
unsuccessful. Full-ratio re-conditioning further transfers these discoveries
to a unified policy while
requiring no additional mechanism at inference time. Extensive evaluations
across diverse model backbones and video understanding and reasoning benchmarks
demonstrate that FrameGRPO delivers robust overall improvements over standard
GRPO. Together with the greater within-group reward variation observed during
training, these results highlight temporal-view diversification as an effective
approach to exploration and credit assignment for video MLLMs.

\subsection*{AI use statement}


In this work, we used generative AI tools only to assist with drafting and
polishing the manuscript for clarity, grammar, and readability. We did not use
generative AI tools to generate synthetic datasets, develop theoretical models
or conceptual frameworks, formulate or prove mathematical claims, propose or
refine hypotheses, design the research methodology or experiments, implement
methods, translate content, clean or reformat datasets, conduct qualitative or
thematic data analysis, or interpret results. All AI-assisted text was manually
reviewed, revised, and verified by the authors for accuracy, originality, and
consistency with the research. We take full responsibility for the final content
of this work, including any text produced with the aid of generative AI.

\bibliography{iclr2027_conference}
\bibliographystyle{iclr2027_conference}

\appendix
\section{Additional Ablation Studies And Results}

\paragraph{Robustness Across Random Seeds.}
To evaluate robustness to training stochasticity, we repeat GRPO and FrameGRPO
with three random seeds on Qwen2.5-VL-7B while keeping all other settings fixed.
As shown in Table~\ref{tab:seed_robustness}, FrameGRPO consistently outperforms
the corresponding GRPO baseline across nearly the entire evaluation suite for
every seed. This consistent trend indicates that the benefits of multi-frame
exploration and gated temporal credit are robust to initialization and data
ordering rather than arising from a favorable individual run.

\begin{table}[H]
\caption{\textbf{Robustness Across Random Seeds.} Results for GRPO and FrameGRPO with Qwen2.5-VL-7B under three random seeds. All results are reported as accuracy (\%). The best result within each seed is highlighted in bold.}
\label{tab:seed_robustness}
\centering
\small
\setlength{\tabcolsep}{1.4pt}
\renewcommand{\arraystretch}{1.15}
\resizebox{\textwidth}{!}{%
\begin{tabular}{l|c|ccc|cccc}
\hline
\multicolumn{2}{c|}{{\footnotesize\textbf{Configuration}}} & \multicolumn{3}{c|}{{\footnotesize\textbf{Video General Benchmark}}} & \multicolumn{4}{c}{{\footnotesize\textbf{Video Reasoning Benchmark}}} \\ \cline{1-9}
{\footnotesize\textbf{Method}} & {\footnotesize\textbf{Seed}} & {\footnotesize\textbf{Video-MME}} & {\footnotesize\textbf{TempCompass}} & {\footnotesize\textbf{VideoEval-Pro}} & {\footnotesize\textbf{MMVU}} & {\footnotesize\textbf{VSI-Bench}} & {\footnotesize\textbf{NExT-GQA}} & {\footnotesize\textbf{Video-MMMU}} \\ \hline
\begin{tabular}{@{}l@{}}GRPO \\ FrameGRPO\end{tabular} & 42 & \begin{tabular}{@{}c@{}}60.0 \\ \textbf{61.3}\end{tabular} & \begin{tabular}{@{}c@{}}74.4 \\ \textbf{74.9}\end{tabular} & \begin{tabular}{@{}c@{}}41.0 \\ \textbf{42.4}\end{tabular} & \begin{tabular}{@{}c@{}}65.6 \\ \textbf{66.7}\end{tabular} & \begin{tabular}{@{}c@{}}34.1 \\ \textbf{37.9}\end{tabular} & \begin{tabular}{@{}c@{}}79.5 \\ \textbf{79.6}\end{tabular} & \begin{tabular}{@{}c@{}}52.0 \\ \textbf{54.2}\end{tabular} \\ \hline
\begin{tabular}{@{}l@{}}GRPO \\ FrameGRPO\end{tabular} & 43 & \begin{tabular}{@{}c@{}}59.8 \\ \textbf{61.3}\end{tabular} & \begin{tabular}{@{}c@{}}73.8 \\ \textbf{75.0}\end{tabular} & \begin{tabular}{@{}c@{}}39.4 \\ \textbf{42.1}\end{tabular} & \begin{tabular}{@{}c@{}}66.4 \\ \textbf{68.0}\end{tabular} & \begin{tabular}{@{}c@{}}34.0 \\ \textbf{35.1}\end{tabular} & \begin{tabular}{@{}c@{}}\textbf{80.0} \\ 79.8\end{tabular} & \begin{tabular}{@{}c@{}}50.2 \\ \textbf{51.4}\end{tabular} \\ \hline
\begin{tabular}{@{}l@{}}GRPO \\ FrameGRPO\end{tabular} & 44 & \begin{tabular}{@{}c@{}}60.2 \\ \textbf{61.5}\end{tabular} & \begin{tabular}{@{}c@{}}73.8 \\ \textbf{74.9}\end{tabular} & \begin{tabular}{@{}c@{}}41.8 \\ \textbf{42.2}\end{tabular} & \begin{tabular}{@{}c@{}}65.6 \\ \textbf{67.7}\end{tabular} & \begin{tabular}{@{}c@{}}31.7 \\ \textbf{34.3}\end{tabular} & \begin{tabular}{@{}c@{}}79.4 \\ \textbf{79.6}\end{tabular} & \begin{tabular}{@{}c@{}}52.4 \\ \textbf{52.7}\end{tabular} \\ \hline
\end{tabular}%
}
\vspace{-3mm}
\end{table}

\paragraph{Scalability Analysis.}
We further compare GRPO and FrameGRPO using training subsets of different
sizes while keeping the remaining settings fixed. As shown in
Table~\ref{tab:data_size_robustness}, FrameGRPO consistently improves the
corresponding GRPO baseline across nearly all evaluation settings, from the
smallest subset to the full training set. The gains remain evident in the
low-data regime and become more broadly consistent as additional training data
are introduced, indicating that FrameGRPO is effective across different data
budgets and can make productive use of increased supervision.

\begin{table}[H]
\caption{\textbf{Robustness Across Training-Set Sizes.} Results for GRPO and FrameGRPO with Qwen2.5-VL-7B using different numbers of training examples. All results are reported as accuracy (\%). The best result within each data size is highlighted in bold.}
\label{tab:data_size_robustness}
\centering
\small
\setlength{\tabcolsep}{1.4pt}
\renewcommand{\arraystretch}{1.15}
\resizebox{\textwidth}{!}{%
\begin{tabular}{l|c|ccc|cccc}
\hline
\multicolumn{2}{c|}{{\footnotesize\textbf{Configuration}}} & \multicolumn{3}{c|}{{\footnotesize\textbf{Video General Benchmark}}} & \multicolumn{4}{c}{{\footnotesize\textbf{Video Reasoning Benchmark}}} \\ \cline{1-9}
{\footnotesize\textbf{Method}} & {\footnotesize\textbf{Data Size}} & {\footnotesize\textbf{Video-MME}} & {\footnotesize\textbf{TempCompass}} & {\footnotesize\textbf{VideoEval-Pro}} & {\footnotesize\textbf{MMVU}} & {\footnotesize\textbf{VSI-Bench}} & {\footnotesize\textbf{NExT-GQA}} & {\footnotesize\textbf{Video-MMMU}} \\ \hline
\begin{tabular}{@{}l@{}}GRPO \\ FrameGRPO\end{tabular} & 2K & \begin{tabular}{@{}c@{}}59.6 \\ \textbf{60.6}\end{tabular} & \begin{tabular}{@{}c@{}}73.7 \\ \textbf{74.0}\end{tabular} & \begin{tabular}{@{}c@{}}41.9 \\ \textbf{42.6}\end{tabular} & \begin{tabular}{@{}c@{}}66.2 \\ \textbf{66.9}\end{tabular} & \begin{tabular}{@{}c@{}}31.0 \\ \textbf{33.1}\end{tabular} & \begin{tabular}{@{}c@{}}\textbf{79.7} \\ 78.9\end{tabular} & \begin{tabular}{@{}c@{}}51.6 \\ \textbf{51.7}\end{tabular} \\ \hline
\begin{tabular}{@{}l@{}}GRPO \\ FrameGRPO\end{tabular} & 4K & \begin{tabular}{@{}c@{}}60.6 \\ \textbf{61.4}\end{tabular} & \begin{tabular}{@{}c@{}}74.0 \\ \textbf{75.2}\end{tabular} & \begin{tabular}{@{}c@{}}42.5 \\ \textbf{42.6}\end{tabular} & \begin{tabular}{@{}c@{}}65.3 \\ \textbf{66.7}\end{tabular} & \begin{tabular}{@{}c@{}}33.4 \\ \textbf{37.6}\end{tabular} & \begin{tabular}{@{}c@{}}\textbf{79.6} \\ 79.5\end{tabular} & \begin{tabular}{@{}c@{}}51.2 \\ \textbf{52.4}\end{tabular} \\ \hline
\begin{tabular}{@{}l@{}}GRPO \\ FrameGRPO\end{tabular} & 8K & \begin{tabular}{@{}c@{}}60.0 \\ \textbf{61.3}\end{tabular} & \begin{tabular}{@{}c@{}}74.4 \\ \textbf{74.9}\end{tabular} & \begin{tabular}{@{}c@{}}41.0 \\ \textbf{42.4}\end{tabular} & \begin{tabular}{@{}c@{}}65.6 \\ \textbf{66.7}\end{tabular} & \begin{tabular}{@{}c@{}}34.1 \\ \textbf{37.9}\end{tabular} & \begin{tabular}{@{}c@{}}79.5 \\ \textbf{79.6}\end{tabular} & \begin{tabular}{@{}c@{}}52.0 \\ \textbf{54.2}\end{tabular} \\ \hline
\end{tabular}%
}
\vspace{-3mm}
\end{table}

\paragraph{Temporal Efficiency Credit Design.}
To isolate the importance of ratio-aware weighting and selective gating in
Temporal Efficiency Credit (TEC), we compare a uniform credit with a fixed
weight, ratio-aware TEC applied without the full-ratio gate, and our gated TEC
under the same multi-frame rollout setting. As shown in
Table~\ref{tab:credit_design_ablation}, assigning a uniform bonus or applying
ratio-aware credit without gating does not yield consistent improvements across
the evaluation suite. A uniform bonus ignores both the amount of temporal
evidence used by a trajectory and whether its success provides information
beyond the full-ratio reference, whereas ungated TEC introduces an unconditional
preference for successful lower-ratio trajectories even when the full-ratio
rollouts already provide valid learning signals. Our gated design achieves the
strongest and most consistent overall performance by preserving the standard
group-relative objective whenever the full-ratio reference succeeds and
activating additional credit only when a lower-ratio trajectory reveals a
complementary success. This comparison confirms that the benefit of TEC arises
not merely from favoring temporally efficient trajectories, but from selectively
crediting them when they contribute otherwise undiscovered reasoning evidence.

\begin{table}[H]
\caption{\textbf{Ablation Study of Temporal Efficiency Credit.} We compare a uniform credit with a fixed weight of $0.2$, TEC without the full-ratio gate, and the proposed gated TEC. All results are reported as accuracy (\%). The best result for each benchmark is highlighted in bold.}
\label{tab:credit_design_ablation}
\centering
\small
\setlength{\tabcolsep}{1.4pt}
\renewcommand{\arraystretch}{1.2}
\resizebox{\textwidth}{!}{%
\begin{tabular}{l|ccc|cccc}
\hline
\raisebox{-5.5pt}[0pt][0pt]{{\footnotesize\textbf{Credit Design}}} & \multicolumn{3}{c|}{{\footnotesize\textbf{Video General Benchmark}}} & \multicolumn{4}{c}{{\footnotesize\textbf{Video Reasoning Benchmark}}} \\ \cline{2-8}
 & {\footnotesize\textbf{Video-MME}} & {\footnotesize\textbf{TempCompass}} & {\footnotesize\textbf{VideoEval-Pro}} & {\footnotesize\textbf{MMVU}} & {\footnotesize\textbf{VSI-Bench}} & {\footnotesize\textbf{NExT-GQA}} & {\footnotesize\textbf{Video-MMMU}} \\ \hline
Uniform Credit ($0.2$) & 60.7 & \textbf{74.9} & 42.0 & 65.1 & 34.7 & 79.3 & 51.3 \\
TEC w/o Gate & 60.1 & 74.6 & 41.1 & 65.9 & 34.9 & 79.5 & 51.9 \\
\rowcolor{blue!6}Gated TEC (Ours) & \textbf{61.3} & \textbf{74.9} & \textbf{42.4} & \textbf{66.7} & \textbf{37.9} & \textbf{79.6} & \textbf{54.2} \\ \hline
\end{tabular}%
}
\vspace{-3mm}
\end{table}

\paragraph{Full-Ratio Reference Subset Size.}
We study the effect of the number of full-ratio reference rollouts $K_\star$
while keeping the group size fixed. As shown in
Table~\ref{tab:reference_subset_ablation}, using a single reference rollout
makes the gate overly sensitive to the outcome of one stochastic generation,
leading to a less reliable estimate of whether the full-ratio view has failed.
Conversely, a large reference subset reserves too many rollout slots for the
same full-ratio view and restricts exploration across alternative temporal
contexts. We therefore adopt $K_\star=2$, which provides a more reliable gating
signal while retaining sufficient rollout capacity for temporal exploration.

\begin{table}[H]
\caption{\textbf{Ablation Study of Full-Ratio Reference Subset Size.} We vary $K_\star$ while keeping the group size fixed. All results are reported as accuracy (\%). The best result for each benchmark is highlighted in bold, and the configuration used in our main experiments is highlighted in blue.}
\label{tab:reference_subset_ablation}
\centering
\small
\setlength{\tabcolsep}{1.4pt}
\renewcommand{\arraystretch}{1.2}
\resizebox{\textwidth}{!}{%
\begin{tabular}{c|ccc|cccc}
\hline
\raisebox{-5.5pt}[0pt][0pt]{{\footnotesize\textbf{Subset Size}}} & \multicolumn{3}{c|}{{\footnotesize\textbf{Video General Benchmark}}} & \multicolumn{4}{c}{{\footnotesize\textbf{Video Reasoning Benchmark}}} \\ \cline{2-8}
 & {\footnotesize\textbf{Video-MME}} & {\footnotesize\textbf{TempCompass}} & {\footnotesize\textbf{VideoEval-Pro}} & {\footnotesize\textbf{MMVU}} & {\footnotesize\textbf{VSI-Bench}} & {\footnotesize\textbf{NExT-GQA}} & {\footnotesize\textbf{Video-MMMU}} \\ \hline
$K_\star=1$ & 60.7 & 74.6 & 41.1 & 65.1 & 35.3 & 78.0 & 49.3 \\
\rowcolor{blue!6}$K_\star=2$ & \textbf{61.3} & 74.9 & \textbf{42.4} & \textbf{66.7} & \textbf{37.9} & \textbf{79.6} & \textbf{54.2} \\
$K_\star=4$ & 60.6 & \textbf{75.0} & 41.4 & 65.0 & 36.6 & 78.9 & 53.4 \\ \hline
\end{tabular}%
}
\vspace{-3mm}
\end{table}

\paragraph{Training Efficiency.}
We compare the end-to-end training cost of GRPO and FrameGRPO in
Table~\ref{tab:training_efficiency}. FrameGRPO maintains comparable training
cost on Qwen2.5-VL-7B and incurs only moderate runtime overhead on Qwen2-VL-7B,
while substantially reducing peak GPU memory for both backbones. These results
show that multi-frame exploration and gated credit assignment can be integrated
without increasing memory demand and with limited additional computation.

\begin{table}[H]
\caption{\textbf{Training Efficiency Comparison.} All runs use eight NVIDIA H20 GPUs. GPU-hours are aggregated across all GPUs.}
\label{tab:training_efficiency}
\centering
\small
\setlength{\tabcolsep}{4pt}
\renewcommand{\arraystretch}{1.08}
\resizebox{0.62\textwidth}{!}{%
\begin{tabular}{lccc}
\hline
\textbf{Method} & \textbf{Wall-Clock Time} & \textbf{GPU-Hours} & \textbf{Peak GPU Memory} \\ \hline
\rowcolor{yellow!10}\multicolumn{4}{c}{\textbf{Qwen2.5-VL-7B}} \\ \hline
GRPO & 21h 24min & 171.2 & 94.95 GiB \\
\rowcolor{blue!6}FrameGRPO & 21h 19min & 170.5 & 57.46 GiB \\ \hline
\rowcolor{yellow!10}\multicolumn{4}{c}{\textbf{Qwen2-VL-7B}} \\ \hline
GRPO & 17h 01min & 136.1 & 94.63 GiB \\
\rowcolor{blue!6}FrameGRPO & 19h 51min & 158.8 & 63.80 GiB \\ \hline
\end{tabular}%
}
\vspace{-3mm}
\end{table}

\section{More Implementation Details}

\subsection{Training Details}
Table~\ref{tab:training_hyperparameters} summarizes the training configuration
used in our FrameGRPO experiments.

\begin{table}[H]
\caption{\textbf{Training Hyperparameters for FrameGRPO.} The table reports the default configuration used in our experiments.}
\label{tab:training_hyperparameters}
\centering
\small
\setlength{\tabcolsep}{8pt}
\renewcommand{\arraystretch}{1.08}
\begin{tabular}{p{0.48\textwidth}p{0.42\textwidth}}
\hline
\textbf{Hyperparameter} & \textbf{Value} \\ \hline
\multicolumn{2}{l}{\textit{Sampling}} \\
Group size $G$ & 8 rollouts per prompt \\
Training sampling temperature & 1.0 \\
Maximum prompt / completion length & 8,192 / 768 tokens \\ \hline
\multicolumn{2}{l}{\textit{Optimization}} \\
Optimizer & AdamW \\
Learning rate & $1\times10^{-6}$ \\
Gradient clipping norm & 1.0 \\
Training epochs / optimizer steps & 1 / 1,000 \\
Prompt batch size per GPU & 1 \\ \hline
\multicolumn{2}{l}{\textit{GRPO / FrameGRPO}} \\
KL coefficient $\beta$ & 0.04 \\
Full-ratio frame count $f_\star$ & 32 \\
Full-ratio reference subset size $K_\star$ & 2 \\
TEC coefficient $\alpha$ & 0.2 \\\hline
\end{tabular}
\vspace{-3mm}
\end{table}

\subsection{Prompt Template}
\begin{figure}[H]
    \centering
    \includegraphics[width=\linewidth]{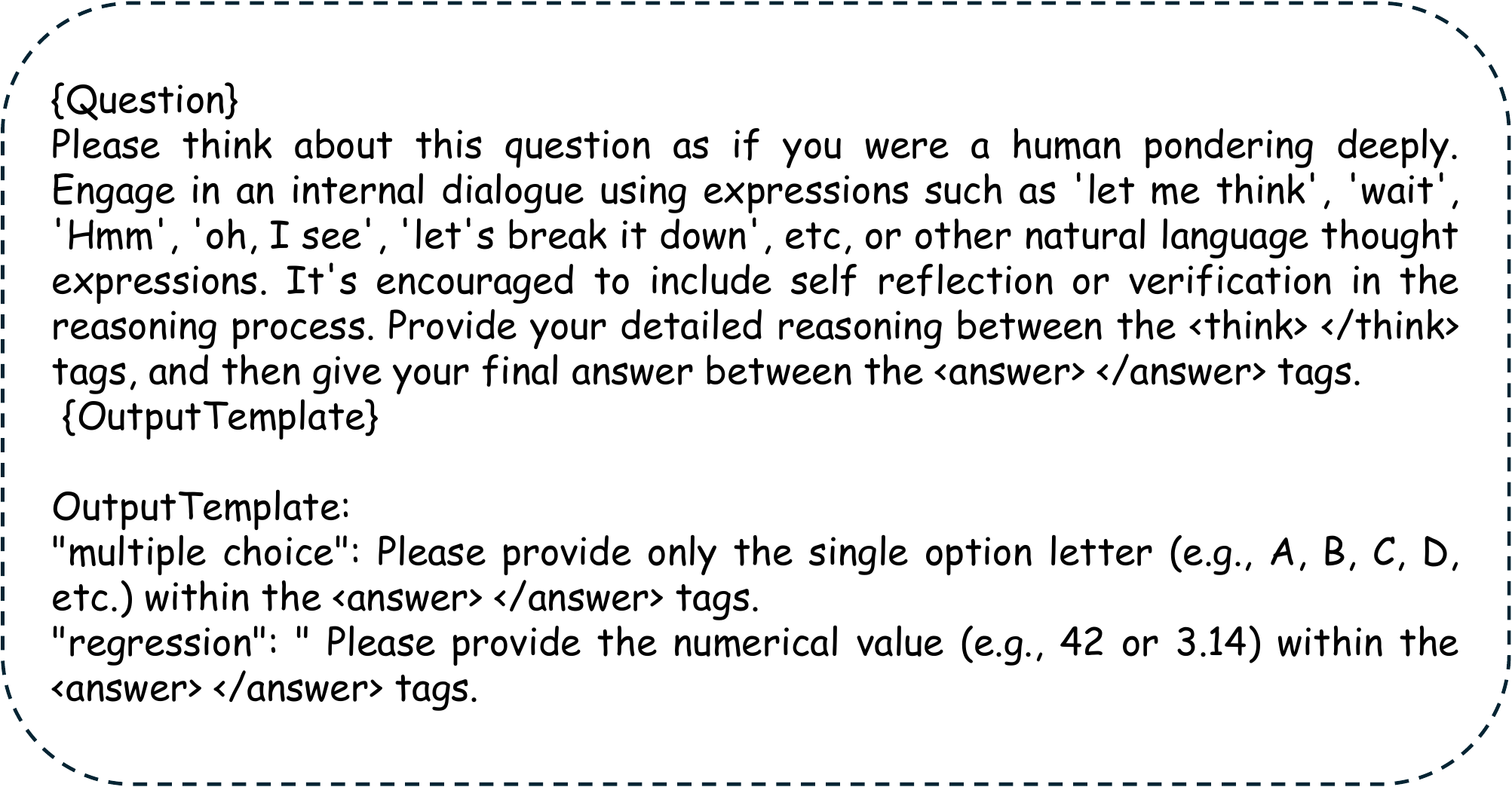}
    \caption{\textbf{Prompt template used for both training and inference.}}
    \label{fig:prompt-template}
\end{figure}

We use the same prompt template for both training and inference, as shown in
Figure~\ref{fig:prompt-template}. The question and task-specific output format
are inserted into the corresponding placeholders, ensuring that the reasoning
instructions and answer format remain consistent across optimization and
evaluation.

\section{Reasoning Example}
\begin{figure}[H]
    \centering
    \includegraphics[width=\linewidth]{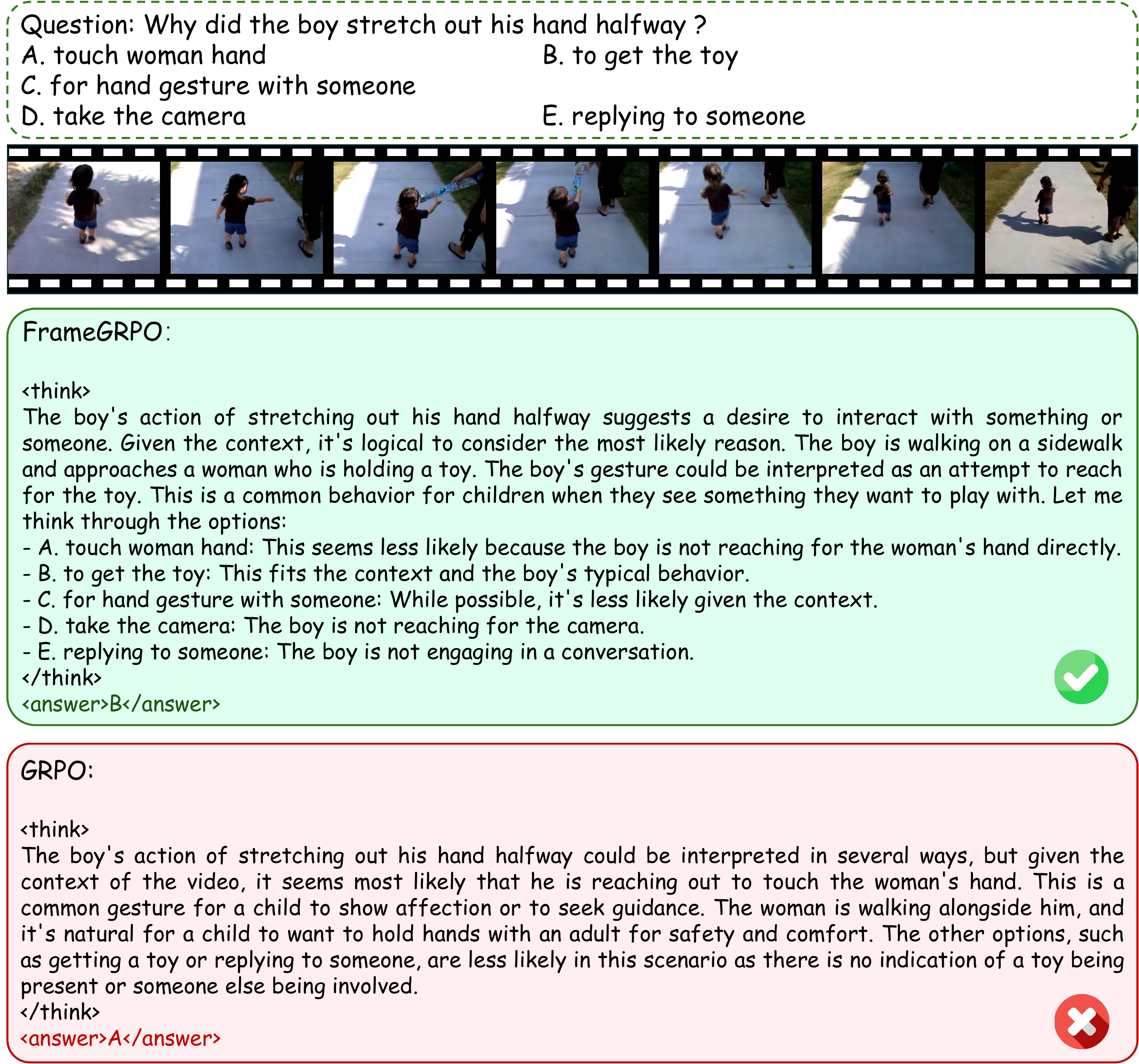}
    \caption{\textbf{Reasoning example on NExT-GQA.}}
    \label{fig:nextgqa-reasoning-example}
\end{figure}

\begin{figure}[H]
    \centering
    \includegraphics[width=\linewidth]{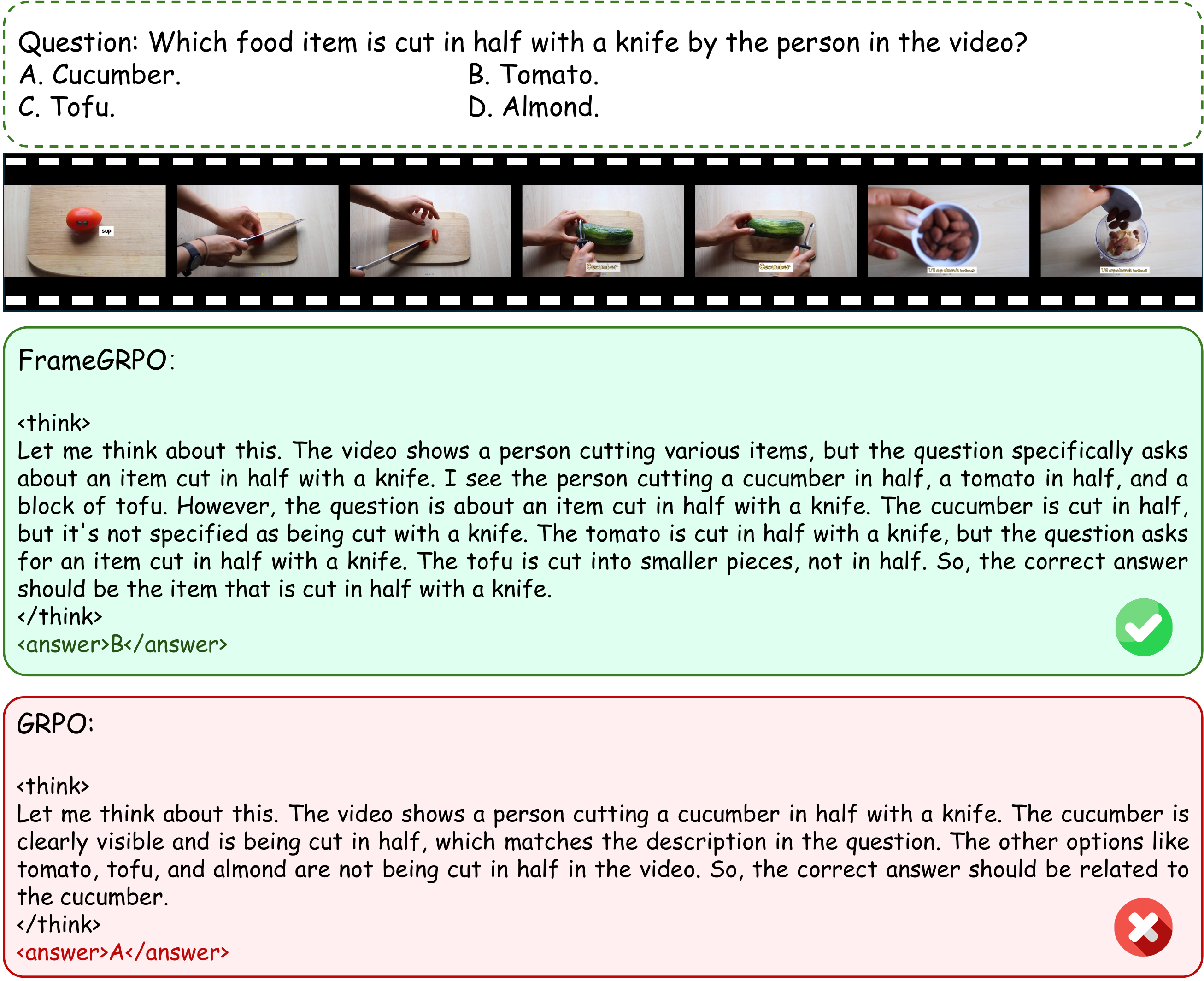}
    \caption{\textbf{Reasoning example on Video-MME.}}
    \label{fig:videomme-reasoning-example}
\end{figure}

We provide two representative case studies comparing FrameGRPO with vanilla
GRPO to highlight the improvements in video perception and reasoning, as shown
in Figures~\ref{fig:nextgqa-reasoning-example} and
\ref{fig:videomme-reasoning-example}, respectively.

\clearpage
\vspace*{\fill}
\vspace*{\fill}
\clearpage

\vspace*{\fill}
\vspace*{\fill}

\end{document}